\documentclass{article} 
\usepackage{iclr2021_conference,times}

\usepackage{amsmath,amsfonts,bm}

\def\eqref#1{equation~\ref{#1}}

\def\1{\bm{1}}

\DeclareMathAlphabet{\mathsfit}{\encodingdefault}{\sfdefault}{m}{sl}
\SetMathAlphabet{\mathsfit}{bold}{\encodingdefault}{\sfdefault}{bx}{n}

\DeclareMathOperator*{\argmax}{arg\,max}

\usepackage{hyperref}
\usepackage{url}

\usepackage{booktabs, multirow} 
\usepackage{soul}
\usepackage{pifont}
\usepackage{graphicx}
\usepackage{array}
\usepackage{amssymb}
\usepackage{changepage,threeparttable}
\usepackage{longtable}
\usepackage{xcolor} 
\usepackage{hyperref}

\definecolor{RoyalBlue}{RGB}{65,105,225}
\definecolor{Teal}{RGB}{0,128,128}
\definecolor{DarkGreen}{RGB}{0,100,0}
\definecolor{DarkOrchid}{RGB}{153,50,204}

\hypersetup{ 
    colorlinks=true, 
    linkcolor=RoyalBlue,
    filecolor=RoyalBlue,
    citecolor=RoyalBlue,
    urlcolor=RoyalBlue,
}

\title{\includegraphics[width=0.6cm]{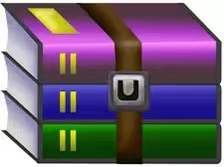} RAR-b: Reasoning as Retrieval Benchmark}

\author{Chenghao Xiao \quad G Thomas Hudson\quad Noura Al Moubayed \\
Department of Computer Science\\
Durham University\\
\texttt{\{chenghao.xiao,g.t.hudson,noura.al-moubayed\}@durham.ac.uk} \\
}

\iclrfinalcopy 
\begin{document}

\maketitle

\begin{abstract}

Semantic textual similartiy (STS) and information retrieval tasks (IR) tasks have been the two major avenues to record the progress of embedding models in the past few years. Under the emerging Retrieval-augmented Generation (RAG) paradigm, we envision the need to evaluate next-level language understanding abilities of embedding models, and take a conscious look at the reasoning abilities stored in them. Addressing this, we pose the question: \textbf{Can retrievers solve reasoning problems?} By transforming reasoning tasks into retrieval tasks, we find that without specifically trained for reasoning-level language understanding, current state-of-the-art retriever models may still be far from being competent for playing the role of assisting LLMs, especially in reasoning-intensive tasks. Moreover, albeit trained to be aware of instructions, instruction-aware IR models are often better off without instructions in inference time for reasoning tasks, posing an overlooked retriever-LLM behavioral gap for the research community to align. However, recent decoder-based embedding models show great promise in narrowing the gap, highlighting the pathway for embedding models to achieve reasoning-level language understanding. We also show that, although current off-the-shelf re-ranker models fail on these tasks, injecting reasoning abilities into them through fine-tuning still appears easier than doing so to bi-encoders, and we are able to achieve state-of-the-art performance across all tasks by fine-tuning a reranking model. We release Reasoning as Retrieval Benchmark (RAR-b), a holistic suite of tasks and settings to evaluate the reasoning abilities stored in retriever models.
\textbf{RAR-b} is available at \url{https://github.com/gowitheflow-1998/RAR-b}.
\end{abstract}

\section{Introduction}

Semantic textual similarity (STS) and information retrieval tasks (IR) have been two principle measures to record the progress of dense representation models \citep{agirre2013sem,agirre2014semeval,agirre2015semeval,agirre2016semeval,cer2017semeval,thakur2021beir}. Despite still heavily being evaluated in sentence representation research, STS is known for its limited alignment with real-world use cases \citep{neelakantan2022text,muennighoff2023mteb}, ambiguity \citep{deshpande2023csts}, and performance orthogonality with IR and other downstream tasks \citep{reimers2016task,wang2021tsdae,xiao2023length}. 

In the LLM era, Retrieval-augmented Generation (RAG) \citep{lewis2020retrieval,neelakantan2022text,xu2023retrieval,gao2023retrieval} has become a go-to alternative method to vanilla end-to-end generative language models \citep{gpt4,touvron2023llama}. This shift is motivated by the inherent weaknesses of LLMs towards factual errors, due to hallucinations \citep{ji2023survey}, knowledge outdatedness \citep{vu2023freshllms}, rarity in long-tailed knowledge \citep{kandpal2023large,malaviya2023expertqa}, and reasoning failure such as on logical deduction \citep{berglund2023reversal}.

Retrieval-Augmented Generation (RAG) is employed differently across various NLP tasks:

\begin{itemize}
    \item For knowledge-intensive tasks, RAG is employed to retrieve the most up-to-date and reliable knowledge references \citep{vu2023freshllms, malaviya2023expertqa}, serving as new prompts for LLMs to extract information and formulate responses. This method mitigates models' natural tendencies to hallucinate \citep{ji2023survey} and reduces the need for frequent fine-tuning of LLMs.
    \item In reasoning-dependent tasks, RAG aims to fetch the most relevant chunks from extensive inputs to guide the focus of LLMs, e.g., in multi-hop question answering scenarios where reasoning across chunks from multiple documents is required. Reasoning with such long context is not only impossible for LLMs with built-in short context windows, but also challenging for LLMs with long context capabilities \citep{xu2023retrieval}.
\end{itemize}

Despite the promise shown by dense retrievers in fetching references for knowledge-intensive tasks, these systems still fall short in retrieving reliable and cite-worthy references \citep{malaviya2023expertqa}, compared to state-of-the-art proprietary LLMs \citep{gpt4} in a standalone manner, highlighting the undesirable behavior of retrievers in assisting LLMs \citep{behnamghader-etal-2023-retriever,asai2024reliable}. This discrepancy is more pronounced in reasoning-intensive tasks, where retrieval-augmented generation methods present inconsistent gains, or even performance degradation to LLMs \citep{bai2023longbench,xu2023retrieval}. 


With the complexities of the role that dense representation models play in the LLM era, the need for accurately assessing their advanced language understanding abilities becomes crucial. We advocate for evaluating these models' capabilities beyond mere factual recall or semantic matching, focusing on their proficiency in complex thought processes and logical deduction.

This paper introduces the Reasoning as Retrieval Benchmark (RAR-b), a novel framework that reframes reasoning tasks as retrieval tasks, offering a comprehensive reevaluation of \textbf{``the actual reasoning representation compressed"}\footnote{we use \includegraphics[width=0.3cm]{figures/rar.png} ``RAR" to convey the connotation of ``compression"} in dense representation models, and challenges the status quo of reasoning-oriented RAG.

Historically, lexical-based retrieval methods have long been seen as baselines of reasoning tasks \citep{clark2016combining,rogers2020getting,sakaguchi2021winogrande}, and often proved insufficient \citep{clark2018think}. For instance, the ARC challenge \citep{clark2018think} itself is constructed by filtering the examples where retrieval systems fail. With the advancements of dense retrieval systems and the increased adoption of RAG, RAR-b emerges as a timely and crucial framework, prompting the essential question: Can dense representation models effectively encode and utilize the reasoning abilities necessary for complex language understanding?
 

RAR-b's contributions are three-fold:

\begin{itemize}
    \item With the established consensus of semantic similarity understanding ability \citep{cer2017semeval} and topical alignment ability \citep{thakur2021beir} possessed by sentence and document representation models, RAR-b takes a step further and envisions their possession of \textbf{reasoning abilities}, and calls for evaluating them today.
    \item We extensively evaluate state-of-the-art open-source and proprietary retrievers and re-rankers, covering unsupervised, supervised, instruction-aware ones, providing an in-depth understanding of reasoning behaviors of current retrieval models.
    \item We reveal key insights through comparing behaviors yielded by systems with different architectures and training recipes, highlighting pathways for embedding models to achieve reasoning-level language understanding and directing future research in the field.
\end{itemize}

\section{RAR-b}

In this work, we construct and release \textbf{RAR-b}: Reasoning as Retrieval Benchmark. Deviating from evaluating on reasoning tasks with a full RAG pipeline (retriever+LLM), we instead focus on evaluating only the retrievers. By constructing reasoning tasks into retrieval tasks, we investigate how good retrievers are on solving them in a standalone manner, and use this as a proxy of the upperbound of retrievers' capabilities in assisting LLMs, in a standard RAG system.

We design three levels of tasks, resulting in the integration of 12 tasks derived from 17 datasets. We convert the original datasets into retrieval formats with both multiple-choice retrieval and full-dataset retrieval settings. 
We first benchmark the performance of state-of-the-art bi-encoder models,
spanning across three model categories: unsupervised dense retrieval models, supervised dense retrieval models, and instruction-aware dense retrieval models. We evaluate both representative open-source models and proprietary models such as Cohere and OpenAI Embedding Models. We further benchmark the performance of representative re-ranking models, both on using them to solve multiple-choice retrieval setting independently, and on
further re-ranking the documents retrieved by bi-encoders in the Full-dataset retrieval setting.

\subsection{Problem Formulation}

We propose the novel framework of evaluating reasoning tasks as retrieval tasks, assessing whether representations of reasoning expressions are semantically well-encoded by retrievers. In other words, we assess whether the groundtruth answer to a reasoning problem is encoded to have the highest similarity with the query in the semantic space. Given a retriever $R$, and a query $q \in Q$, and a groundtruth answer $g$ hidden in a corpus $\mathcal{C}$ that is intentionally made very large, the most ideal scenario would be:
\begin{equation} 
\argmax_{d \in C} S(R(q), R(d)) = g
\end{equation}
However, this is typically not possible given the complexity of reasoning language and the fact that current sentence encoders are typically not yet well-trained to model such expressions. Therefore, we in turn are interested in whether:
\begin{equation}  g \in \argmax\text{}^{(n)}_{d \in C} S(R(q), R(d))
\end{equation}
For the Full-dataset retrieval setting, we can quantitatively measure $R$'s performance with commonly-adopted information retrieval metrics such as nDCG@n and recall@n, respectively concerning how well the top-n retrieved answers are ranked (nDCG@n), and if the correct answers are even retrieved as the top n (recall@n).

For the Multiple-choice setting, we simply assess whether the query is perceived by the retrievers to be more similar to the groundtruth answer, than to other candidate answers in its original dataset format. Models' performance in this setting can be easily quantified by accuracy.

\subsection{Datasets}

\paragraph{Dataset and Processing}
The datasets we consider can all be seen as reasoning problems. Firstly, we include commonsense reasoning datasets, $\alpha$NLI \citep{bhagavatula2019abductive}, HellaSwag \citep{zellers2019hellaswag}, PIQA \citep{bisk2020piqa}, SocialiQA \citep{sap2019social}, ARC-Challenge \citep{clark2018think}, Quail \citep{rogers2020getting}, Winogrande \citep{sakaguchi2021winogrande}, and C-STS \citep{deshpande2023csts}. For all datasets in the full retrieval settings, we pool all the candidate answers across all splits to serve as the corpus, aligning with the setting in most retrieval benchmarks, and evaluate them using test set questions as the queries. All commonsense reasoning datasets are suitable to test in Full-dataset retrieval setting, except for C-STS \citep{deshpande2023csts} due to the potential sparse annotation problem when constructing it into Full-dataset retrieval settings \citep{thakur2021beir} (detailed explanation in Appendix~\ref{appendix: csts sparse annotation}).

Apart from commonsense reasoning abilities, we further envision the possession of temporal, spatial, numerical, and symbolic reasoning abilities in dense retrievers. Temporal and spatial reasoning abilities are evaluated respectively with TempReason \citep{tan2023towards} and SpartQA \citep{mirzaee2021spartqa}. For TempReason, we evaluate each of its sub-level tasks separately, and construct the pure, fact and context settings to assess different aspects of retrievers' behaviors (detailed analyses in Section~\ref{subsec: temp reason}). For numerical reasoning abilities, we concatenate MATH \citep{hendrycks2021measuring} and GSM8K \citep{cobbe2021training}, two datasets commonly used to evaluate LLMs, using questions as queries, the pool of all answers as the corpus. Because of their small scales, we enlarge the corpus with MetaMathQA \citep{yu2023metamath}, which is created synthetically with LLMs, using the training set of MATH. Therefore, it is assured that no examples in MetaMathQA can act as the groundtruth answer for any of our evaluated queries (which are from the test set of GSM8K and MATH). 

With the common user cases of code retrieval and that code understanding can serve as a proxy of the understanding of symbolic language, we further include code retrieval tasks. We concatenate HumanEvalPack \citep{muennighoff2023octopack} and MBPP \citep{austin2021program} to form the evaluation queries, because of their validated quality, ubiquitously seen in the evaluation of LLMs. Notably, HumanEvalPack \citep{muennighoff2023octopack} is an extended version of HumanEval \citep{chen2021evaluating}, by translating the original Python split to cover Javascript, Java, Go, C++, and Rust. To enlarge the corpus, we further sample 200k code from CodeSearchNet \citep{husain2019codesearchnet} and 100k answers from a synthetic dataset TinyCode\footnote{\url{https://huggingface.co/datasets/nampdn-ai/tiny-codes}} to cover pure code text and the mixture of natural language and code. We extensively explore the optimal setting to construct the code corpus and evaluate the code tasks (see analyses in Section~\ref{sec: ablation entity shortcut}) to make it as difficult as possible while keeping the evaluation computationally efficient. For HumanevalPack, we evaluate the code continuation setting; when enlarging the corpus, we sample examples from CodeSearchNet that fall under similar length range of the groudtruth documents to make the retrieval more difficult, given that length information is typically encoded in the embeddings \citep{xiao2023length} and we don't want the models to leverage this information to simplify the retrieval.

\begin{table*}[t!]
    \small
    \resizebox{\textwidth}{!}{\begin{tabular}{ l | l | l | c | c | c | c  c | c c }
        \toprule
         \multicolumn{1}{l}{\textbf{Split} ($\rightarrow$)} &
         \multicolumn{3}{c}{} & 
         \multicolumn{1}{c}{\textbf{Train}}    &
         \multicolumn{1}{c}{\textbf{Dev}}    &
         \multicolumn{2}{c}{\textbf{Test}}   &
         \multicolumn{2}{c}{\textbf{Avg.~\#Words}} \\
         \cmidrule(lr){5-5}
         \cmidrule(lr){6-6}
         \cmidrule(lr){7-8}
         \cmidrule(lr){9-10}
           \textbf{Task ($\downarrow$)} &\textbf{Domain ($\downarrow$)} & \textbf{Dataset ($\downarrow$)} & \textbf{\#MCR} & \textbf{\#Pairs} & \textbf{\#Query} & \textbf{\#Query} & \textbf{\#Corpus$\clubsuit$}& \textbf{Query} & \textbf{Document} \\
         \midrule
    Commonsense &  General & $\alpha$NLI \citep{bhagavatula2019abductive} &  2 & 169654 & 0 ($\spadesuit$) & 1532 &  241347 & 19.35 & 8.30 \\ 
    Commonsense &  General & HellaSwag \citep{zellers2019hellaswag} & 2 & 39905 & 0 ($\spadesuit$) & 10042 & 199162 & 40.12 & 24.71  \\ 
    Commonsense & Physical & PIQA \citep{bisk2020piqa} & 2  & 16113 & 0 ($\spadesuit$)  &1838 & 35542 & 7.08 & 18.90\\
    Commonsense & Social & SIQA \citep{sap2019social} &  3 & 33410 & 0 ($\spadesuit$) & 1954 &  71276 & 22.28  & 4.39  \\
    Commonsense & Multiple & Quail \citep{rogers2020getting}  & 4 & 10246 & 0 ($\dagger$) & 2720 & 32787 & 345.32 & 4.98 \\
    Commonsense & Scientific & ARC-C \citep{clark2018think}  & 4 & 1119 &  299 & 1172 & 9350 & 22.65 & 5.29\\
    Commonsense & General & WinoGrande \citep{sakaguchi2021winogrande}  &  2 & 40398 & 0 ($\spadesuit$)  & 1267  & 5095 & 20.11& 1.22 \\
    Commonsense & General & c-STS \citep{deshpande2023csts}  & 2 &       &   &     & ---$\bigstar$ & &    \\
    \midrule[0.05pt] \midrule[0.05pt]   
   
    Temporal & General & TempReason \citep{tan2023towards}  & ---  &  &  &    & &  &    \\
    Temporal & General & \quad - TR1  & ---  & 400000 & 4000 & 4000 & 12504 & 11.56 & 2.00   \\
    Temporal & General & \quad - TR2  & ---  & 16017 & 5521 & 5397 & 15787 & \hspace{0.3em}10.47$\ddagger$ &2.91\\ 
    Temporal & General & \quad - TR3 & ---  & 13014 &  4437 & 4426 & 15664 & \hspace{0.3em}12.20$\ddagger$ &  2.91  \\ \midrule
    Spatial & General & SpartQA \citep{mirzaee2021spartqa}  &  3 & 22749 & 3579 & 3594 & 1592 & 125.67 & 10.09 \\
    \midrule
    Numerical & Math  & + MATH \citep{hendrycks2021measuring} & ---  & 7500 & 0 &  + 5000 & + 5000 & 31.65 & 84.83 \\
    Numerical & Math & + GSM8K \citep{cobbe2021training}  &  --- & 7473  & 0 & + 1319 & +1319 & 47.25 & 52.78\\ 
    Numerical & Math & + MetaMathQA \citep{yu2023metamath} $\star$  &  --- &---&---  &  ---  & + 383057 & --- &  102.39   \\ 
    Numerical & Math & = math-pooled  &  --- &14973 & 0 & = 6319& = 389376     & 34.91 & 101.99 \\ 
    \midrule
    Symbolic  & Code & + HumanevalPack \citep{muennighoff2023octopack}    &  --- & 0  &  0 & + 984 & + 984 & --- & --- \\ 
    Symbolic  & Code & + MBPP \citep{austin2021program}   &  --- &  0 &  0 & + 500  & + 500 & --- & --- \\ 
    Symbolic  & Code & + CodeSearchNet \citep{husain2019codesearchnet} $\star$  &  --- &  --- &  --- & ---  & + 200000 $\star$& --- & --- \\ 
    Symbolic  & Code & + TinyCode $\star$  &  --- &  --- & --- & ---  & + 100000 $\star$ & --- & --- \\ 
    Symbolic  & Code & = Code-pooled    &  --- &  0 & 0 & = 1484  & = 301484 & --- & --- \\ 
    \bottomrule
    \end{tabular}}
    \vspace{-2mm} 
    \caption{\textbf{Statistics of datasets} of RAR-b. $\spadesuit$: We use the original dev set as the test set, and add the original test set candidates to the corpus if available, as the original test set labels of these datasets are not designed to be publicly available. $\dagger$: We concatenate the validation and "challenge" set as the test set, leaving no dev set. $\clubsuit$: We pool the unique set of candidates across all splits as the corpus where available, i.e., corpus is shared across train, dev, and test set. $\bigstar$: c-STS is not suitable for full-dataset retrieval setting, which is because of the effect of sparse annotation problem if doing so. $\ddagger$: For TR2 and TR3, we construct Pure, Fact, and Context Setting, where the average query lengths are \{10.47, 145.14, 1901.19\}, \{12.20, 157.07, 2132.85\}. Notably, most open-source models are not able to process the context setting at once without loss of information. $\star$: The original MetaMathQA was actually a training set generated from the training set of MATH and GSM8K, but we only use its unique answer set as the corpus, so we do not include non-used statistics here. The same goes to CodeSearchNet and TinyCode in code retrieval, where we respectively sample 200k and 100K to enlarge the code corpus.}
    \label{tab:dataset_stats}
\end{table*}

\paragraph{Task Levels} We group the datasets into three levels. 

Level-1 dataset tasks are by their nature more in-distribution to typical datasets used to train IR models, and makes more sense to be solved even without instructions. For instance, PIQA consists of physical goals and possible solutions, which are similar to IR training sets that are question-answer pairs. HellaSwag and $\alpha$NLI respectively look for the end of an unfinished story, and the middle of a story given the start and the ending, which are both extremely similar to the positive pairs constructed in training of unsupervised dense retrieval models such as Contriever \citep{izacard2022unsupervised}, which samples overlapping spans of the text as positive pairs.

Level-1.5 datasets on the other hand, are more out-of-distribution in terms of task categories, and make less sense to be solved without a \textbf{clearly-stated purpose}. For instance, WinoGrande asks to find the correct answer to fill in an unfinished sentence, which is marked by an underscore. We believe this format is not commonly seen in IR training, and as will be shown, current models struggle to develop a nuanced understanding about this task. We further include a novel task, conditioned-sts (CSTS) \citep{deshpande2023csts}, which concerns different semantic similarities between the same sentence pairs, under different conditions. Both of the tasks make less sense to solve without prepending the query with an instruction, unless the models has seen the format in their training. We position CSTS as the most challenging dataset on this level, as the task is, to the best of our knowledge, not similar to any datasets that would be used in the training of IR models. We structure C-STS in two settings: CSTS-Easy and CSTS-Hard (Appendix~\ref{appendix: dataset format}).

Level-2 datasets include $\{$temporal, numerical, spatial, symbolic$\}$ reasoning. These serve as an extra lens to inspect the abilities that researchers and practitioners do not expect the current generation retrievers to have (with the exception of OpenAI's embedding model class, which are trained on a large portion of code data in the first place \citep{neelakantan2022text} from its earliest versions). These abilities are rarely assessed or considered to be retrievers' necessary capabilities. But we instead envisage that representation models have both a necessity and the capacity to attain these abilities. The reasons are as follows:

Consider the following three examples that might occur as queries to a RAG system. These types of requirements are rarely properly evaluated in current IR tasks and benchmarks. In ``retrieve all records where total sales \textit{$>$ 10}", retrievers are required to have \textbf{numerical understanding}; in ``what are the main arguments in \textit{yesterday's} report?", they need to use their \textbf{temporal perception}; in ``what are popular tourist attractions of the state \textit{next to Florida}?", the retrievers need to have \textbf{spatial knowledge} or \textbf{world knowledge}. We see that these abilities are not stored in current SOTA representation models, and thus relevant problems are at the moment typically solved using external pipelines or methods \citep{wang2023knowledgpt}. For one, one can use an extra agent to help filter the records $>$ 10 in a database system, using meta data, instead of relying this part on retrievers (the RAG pipeline thus complicates to agent - retriever - LLM). On the other hand, one can solve these problems by query rewrites \citep{ma2023query} (e.g., chain-of-thoughts retrieval \citep{yoran2023answering} or inductive retrieval \citep{zhang2023iag}). For instance, one can first use an LLM to understand that the states next to Florida are Georgia and Alabama, and rephrase the query to ``what are popular tourist attractions of Georgia and Alabama?". In this case, the pipeline is also complicated into LLM - retriever - LLM. However, information about time, numbers and space is undoubtedly encode-able, if certain dimensions of the representations are allocated to do so. Out of this reason, we do not think that complicated pipelines, such as ones that go through LLM - retriever - LLM, are eventually necessary. Instead, retrievers can take on more responsibility before LLM in a typical RAG pipeline, if not end-to-end \citep{asai2023self}. In conclusion, we call for the evaluation of level-2 tasks today, as a checkpoint of future essential capabilities of retrieval models.


For all datasets, we evaluate on their test sets, and dev/val sets when test set groundtruth labels are not publicly available, leaving room for the research community to investigate leveraging their training sets in the development of future retrievers with reasoning capabilities.

\subsection{Models}

We first benchmark the performance of state-of-the-art bi-encoders, spanning across unsupervised dense retrieval models (U-DR), supervised dense retrieval models (S-DR), supervised instruction-aware dense retrieval models (I-DR). For U-DR models, we have Contriever \citep{izacard2022unsupervised}. For S-DR models, we include two Sentence Transformers \citep{reimers2019sentence} models that have been the most popular in the last few years, \texttt{all-mpnet-base-v2} and \texttt{all-MiniLM-L6-v2}; Dragon+ \citep{lin2023train}, an S-DR model that is progressively distilled with different views of other SOTA models, and BGE-M3 \citep{chen2024bge}. For I-DR models, we include Instructor \citep{su2022one}, BGE (v1.5) \citep{xiao2023c}, and consider all of their model sizes (base, large, XL for Instructor and small, base, large for BGE) to understand the scaling laws (if there's any) between model size and reasoning abilities. We further include E5-Mistral \citep{wang2023improving}, a latest state-of-the-art instruction-aware embedding model that is finetuned from Mistral \citep{jiang2023mistral} with \{instruction, query, answer\} pairs distilled from GPT-4. We recognize the recent advancements in unifying generative and representation abilities into one model, and benchmark GritLM \citep{muennighoff2024generative}, whose data and training process of the embedding abilities are similar to \cite{jiang2023mistral}, but differing in its full parameters finetuning and joint learning with the generative objective. Lastly, we benchmark popular proprietary models Cohere \texttt{Embed-English-v3.0} and OpenAI \texttt{text-embedding-ada-002}, \texttt{text-embedding-3-small}, \texttt{text-embedding-3-large} through API encoding.

For rerankers (cross-encoders), we benchmark a few representative models such as BGE-reranker-large \citep{xiao2023c}, TART-Full \citep{asai2022task} and Cohere \texttt{rerank-english-v2.0}.

\subsection{Reasoning as Retrieval}

Instead of solving these reasoning tasks via text-to-text generation \citep{raffel2020exploring,lourie2021unicorn}, we explore the possibilities of solving them as a unified representation-to-representation matching problem. That is, all tasks, contexts, and possible answers are all understood through a shared representation space, and the search of the answer, given a task and the context information, falls back to be a simple similarity match problem.

We explore two settings: RAR\textsubscript{w/o Instructions} and RAR\textsubscript{w/ Instructions}. For RAR\textsubscript{w/o Instructions}, we simply use the original question/query/context in the datasets as the query, without describing the task with an instruction prompt, and use the pool of all possible choices across splits as the candidate documents. For RAR\textsubscript{w/ Instructions}, we prepend the query with an instruction that describes the task. For instance, for $\alpha$NLI, an (imperfect) instruction can be ``Given the following start and end of a story, retrieve a possible reason that leads to the end".

We construct two task settings: Multiple-choice Retrieval Setting (MCR), and Full-dataset Retrieval Setting (Full). For multiple choice retrieval, we utilize the choices available to each question in the original dataset. For this setting, the performance can be easily measured by accuracy, as each input has only 2-4 candidates. For Full-dataset Retrieval Setting, we construct a pool of all candidate choices with all the unique candidates from the same dataset, and investigate whether the groundtruth answer can be retrieved from a candidate pool of a much larger order of magnitude. In line with typical IR benchmarks, the corpus is constructed with candidates from all splits, and it is only shared across them in inference. For Full-dataset Retrieval, we measure the performance with commonly-adopted information retrieval metrics, including nDCG@n and Recall@n.

\begin{figure}[t]
    \centering
    \includegraphics[width=1\linewidth]{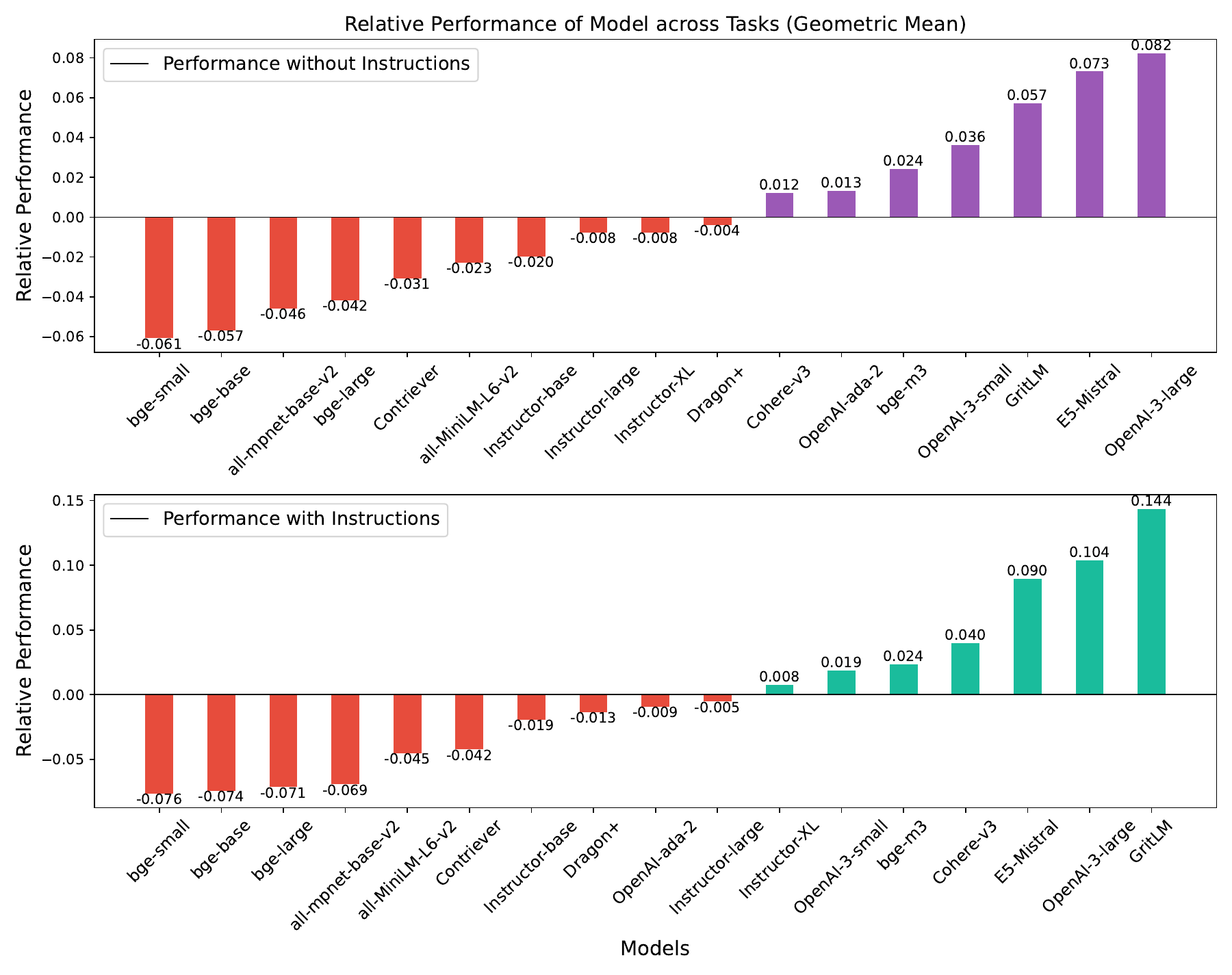}
    \caption{Relative Performance of all models on Full-dataset Retrieval setting. We take the geometric mean across tasks to represent each model's performance; and we subtract the mean performance across models from each model's performance to understand their relative performance.}
    \label{fig: relative performance (main)}
\end{figure}

The multiple-choice setting and the full retrieval setting serve as simulacra of real-life RAG scenarios. Full retrieval performance assesses whether correct answers can be retrieved as top candidates, providing a basis for the later LLMs to make informed decisions. Multiple choice performance gauges whether hard-negative candidates might be erroneously prioritized over the groundtruth answer, introducing noise into the references presented to LLMs. From a pure retrieval perspective on the other hand, multiple-choice setting evaluates the extent to which a dual dense encoder can substitute a cross-encoder in reasoning-intensive late interactions - if dense encoders are already good at MCR, no re-ranking models (cross-encoders) are needed to rerank the top candidates retrieved by dense retrievers, before them being presented to LLMs. As a comparison, we also benchmark the performance of representative re-rankers, both as standalone solves in the MCR setting, and as re-rankers to re-rank candidates retrieved by bi-encoders in the Full setting.

For Multiple-Choice Retrieval, we evaluate both retrievers and re-rankers, on the datasets that are originally multiple-choice problems ($\alpha$NLI, HellaSwag, PIQA, SIQA, WinoGrande), and ones that can be constructed to a MCR problem (C-STS). For Full-Dataset Retrieval, we evaluate both performance of using only retrievers, and retrieval+reranking performance as in a typical retrieval pipeline. We evaluate the FULL setting for every dataset ($\alpha$NLI, HellaSwag, PIQA, SIQA, WinoGrande, TempReason, SpartQA, Math-pooled, Code-pooled) except C-STS, which is not suitable for full retrieval due to potential sparse annotation effect - its original correct answer might not be the most suitable one throughout the corpus.

\section{Results}

\subsection{Full-dataset Retrieval Setting}


\begin{table}[!htp]\centering

\scalebox{0.62}{%
\begin{tabular}{lr|rcccccccccc|cc}\toprule
\multicolumn{2}{l}{Model $\downarrow$ Dataset  $\rightarrow$} & $\alpha$NLI &HellaSwag &PIQA &SIQA &Quail &ARC-C &WinoG  &TR &SpartQA &Math &Code&Geo. Mean\\\midrule

\multirow{2}{*}{Contriever} &w/o Inst. &0.318 &0.144 &0.246 &0.013 &0.050 &0.086 &0.471 &0.108 &0.109 &0.308 &0.093 &0.123 \\
&w/ Inst. &0.271 &0.177 &0.217 &0.009 &0.049 &0.076 &0.263 &0.105 &0.106 &0.218 &0.071 &0.104\\
\midrule
\multirow{2}{*}{all-mpnet-base-v2} &w/o Inst. &0.224 &0.263 &0.290 &0.024 &0.034 &0.118 &0.207 &0.058 &0.002 &0.718 &0.531&0.108 \\
&w/ Inst. &0.020 &0.130 &0.272 &0.013 &0.030 &0.104 &0.097 &0.044 &0.010 &0.692 &0.488&0.077 \\
\multirow{2}{*}{all-MiniLM-L6-v2} &w/o Inst. &0.282 &0.242 &0.253 &0.016 &0.039 &0.095 &0.473 &0.080 &0.017 &0.682 &0.440&0.131\\
&w/ Inst. &0.151 &0.205 &0.247 &0.015 &0.035 &0.094 &0.207 &0.076 &0.006 &0.624 &0.423&0.101 \\
\multirow{2}{*}{Dragon+} &w/o Inst. &0.321 &0.277 &0.280 &0.020 &0.041 &0.089 &0.672 &0.087 &0.103 &0.451 &0.176&0.150 \\
&w/ Inst. &0.252 &0.241 &0.264 &0.017 &0.042 &0.082 &0.609 &0.081 &0.108 &0.362 &0.128 &0.133\\
\multirow{2}{*}{bge-m3} & w/o Inst. & 0.247 & 0.257 & 0.229 & 0.049 & 0.075 & 0.090 & 0.417 & 0.140 & 0.075 & 0.692 & 0.388 &0.178\\
 & w/ Inst. & 0.247 & 0.255 & 0.190 & 0.048 & 0.071 & 0.090 & 0.353 & 0.152 & 0.070 & 0.645 & 0.396 &0.170\\
\midrule
\multirow{2}{*}{Instructor-base} &w/o Inst. &0.246 &0.263 &0.282 &0.024 &0.036 &0.102 &0.301 &0.059 &0.036 &0.584 &0.415 &0.134\\
&w/ Inst. &0.201 &0.239 &0.254 &0.023 &0.038 &0.096 &0.164 &0.052 &0.069 &0.582 &0.401 &0.127\\
\multirow{2}{*}{Instructor-large} &w/o Inst. &0.248 &0.295 &0.332 &0.037 &0.051 &0.125 &0.264 &0.068 &0.016 &0.710 &0.546 &0.146\\
&w/ Inst. &0.234 &0.266 &0.286 &0.031 &0.044 &0.108 &0.226 &0.062 &0.034 &0.672 &0.527 &0.141\\
\multirow{2}{*}{Instructor-XL} &w/o Inst. &0.328 &0.323 &0.364 &0.040 &0.059 &0.144 &0.564 &0.084 &0.003 &0.599 &0.504&0.146\\
&w/ Inst. &0.282 &0.302 &0.319 &0.043 &0.056 &0.115 &0.324 &0.076 &0.022 &0.580 &0.495 &0.154\\
\multirow{2}{*}{bge-small} &w/o Inst. &0.116 &0.254 &0.239 &0.008 &0.017 &0.090 &0.103 &0.077 &0.036 &0.450 &0.424 &0.093\\
&w/ Inst. &0.013 &0.234 &0.208 &0.010 &0.020 &0.077 &0.054 &0.073 &0.029 &0.465 &0.415 &0.070\\
\multirow{2}{*}{bge-base} &w/o Inst. &0.110 &0.266 &0.257 &0.009 &0.014 &0.097 &0.138 &0.076 &0.034 &0.469 &0.465 &0.097\\
&w/ Inst. &0.041 &0.240 &0.230 &0.002 &0.012 &0.088 &0.103 &0.073 &0.027 &0.456 &0.463 &0.072\\
\multirow{2}{*}{bge-large} &w/o Inst. &0.131 &0.285 &0.280 &0.010 &0.018 &0.100 &0.192 &0.111 &0.030 &0.574 &0.481 &0.112\\
&w/ Inst. &0.009 &0.262 &0.233 &0.006 &0.027 &0.089 &0.103 &0.096 &0.023 &0.498 &0.453 &0.075\\

\multirow{2}{*}{E5-Mistral} & w/o Inst.	& 0.189	& 0.322	& 0.328	& 0.051	& 0.070 & 0.205 & 0.452 & 0.185 & 0.109 & 0.779 & 0.798&0.227\\
 & w/ Inst. & 0.261 & 0.349 & 0.394 & 0.054 & 0.081 & 0.178 & 0.412 & 0.188 & 0.099 & 0.740 & 0.785&0.236\\

\multirow{2}{*}{GritLM} & w/o Inst.	& 0.296 & 0.360 & 0.358 & 0.057 & 0.087 & 0.166 & 0.521 & 0.212 & 0.016 & 0.830 & 0.832 & 0.211\\
 & w/ Inst. & 0.340 & 0.395 & 0.444 & 0.072 & 0.116 & 0.266 & 0.537 & 0.268 & 0.094 & 0.824 & 0.838 & 0.290\\
 
\midrule[0.5pt]\midrule[0.5pt]
\multirow{2}{*}{Cohere-Embed-v3} & w/o Inst. & 0.151 & 0.263 & 0.285 & 0.043 & 0.041 & 0.099 & 0.580 & 0.151 & 0.038 & 0.723 & 0.572&0.166\\
 & w/ Inst. & 0.187 & 0.290 & 0.279 & 0.050 & 0.078 & 0.101 & 0.650 & 0.171 & 0.033 & 0.721 & 0.566 &0.186\\

\multirow{2}{*}{OpenAI-ada-002} &w/o Inst. &0.256 &0.293 &0.310 &0.031 &0.058 &0.133 &0.197 & 0.100 &0.042 & 0.732& 0.834 &0.167\\
&w/ Inst. &0.106 &0.248 &0.239 &0.026 &0.058 &0.118 &0.114 &0.096 &0.048 &0.673 & 0.824 &0.137\\

\multirow{2}{*}{OpenAI-3-small} & w/o Inst. & 0.306 & 0.309 & 0.337 & 0.030 & 0.061 & 0.146 & 0.315 & 0.125 & 0.066 & 0.711 & 0.720 &0.190\\
 & w/ Inst. & 0.212 & 0.272 & 0.296 & 0.030 & 0.066 & 0.138 & 0.255 & 0.129 & 0.036 & 0.643 &0.721 &0.165\\

\multirow{2}{*}{OpenAI-3-large} & w/o Inst. & 0.373 & 0.341 & 0.420 & 0.034 & 0.101 & 0.240 & 0.291 & 0.164 & 0.075 & 0.901 & 0.896 &0.236\\
 & w/ Inst. & 0.342 & 0.314 & 0.375 & 0.050 & 0.136 & 0.212 & 0.339 & 0.211 & 0.074 & 0.877 & 0.894 &0.250\\
 
\bottomrule
\end{tabular}}
\caption{Full-dataset Retrieval (nDCG@10 performance)}\label{tab: full-dataset ndcg@10}
\end{table}

\begin{figure}
    \centering
    \includegraphics[width=1\linewidth]{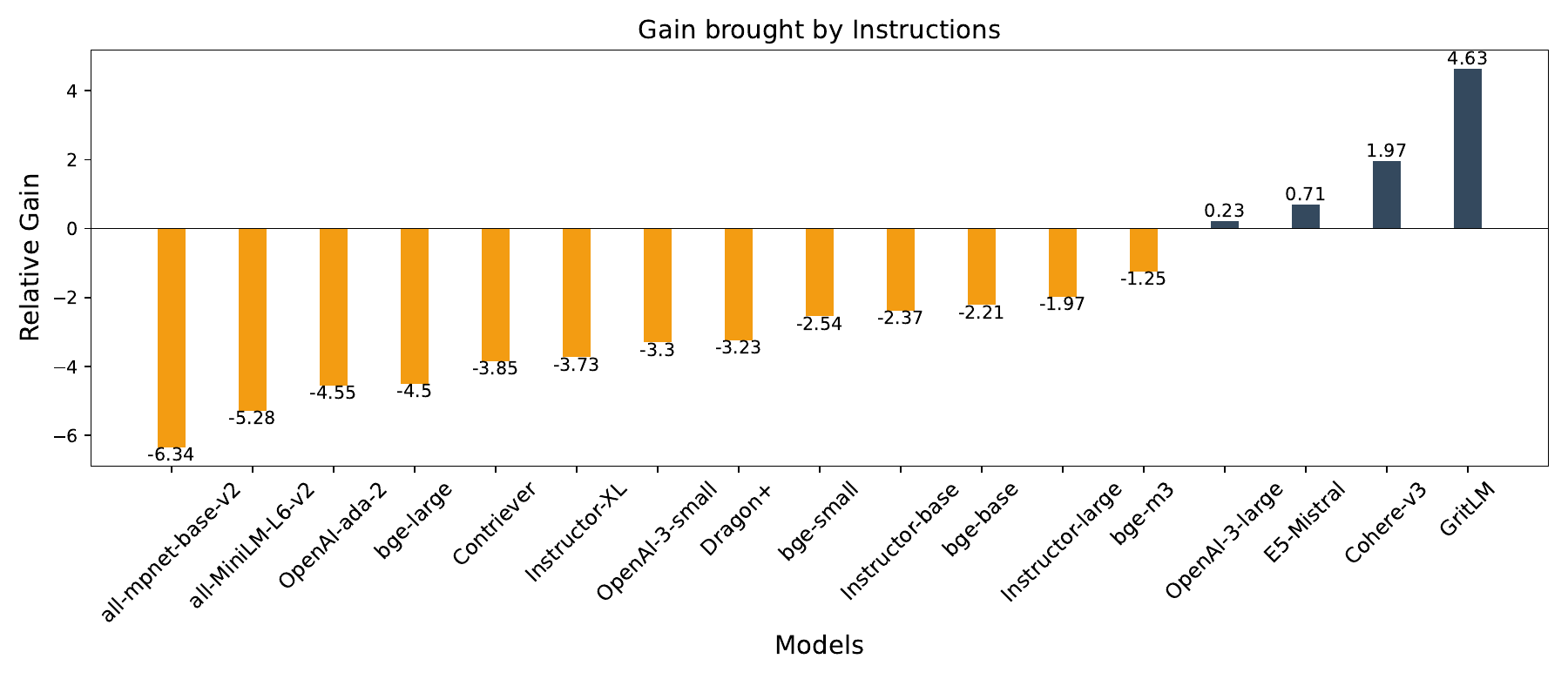}
    \caption{Gain brought by instruction (Full Setting). The metric is represented by log-scaled gains averaged across tasks, given by $\sum_{x \in X}(log_2(x_{\text{w/ inst.}_i}+1)-log_2(x_{\text{w/o inst.}}+1))/|X|$, $X$ being the set of tasks.}
    \label{fig: instruction gain}
\end{figure}

\paragraph{Main results} Table~\ref{tab: full-dataset ndcg@10} presents the results for nDCG@10 performance. Figure~\ref{fig: relative performance (main)} outlines the relative performance of the evaluated models. Because of the different scales of nDCG@10 across tasks due to different task difficulties and corpus sizes, we take the geometric mean across tasks to represent each model's average performance, which is given by $G = \left( \prod_{i=1}^{n} x_i \right)^{\frac{1}{n}}$, where $n$ is the number of tasks and $x_i$ represent the performance of each task. We find geometric mean to be more reflective of model's average performance, compared to harmonic mean and Z-scored mean. Figure~\ref{fig: instruction gain} provides insights into the performance gain/degrade brought by prepending task instructions before queries. Similarly, due to the vastly different scales of performance across tasks, simply taking the arithmetic gain ($\left(x_{\text{with inst.}} - x_{\text{without inst.}} \right)/x_{\text{without inst.}}$) is misleading (by biasing towards a low $x_{\text{without inst.}}$ base). Therefore, we opt for using logarithmic scaling to dampen the effect of large percentage gains from a low base, given by $log_2(x_{\text{with inst.}_i}+1)-log_2(x_{\text{without inst.}}+1)$.

Overall, it is seen that newer models tend to outperform older ones on RAR-b. We believe the enhancement in embedding models' language understanding abilities is relevant to the diversity of training data and instruction-tuning \citep{wang2023improving}, with the current widely-adopted paradigm of instruction-aware information retrieval \citep{asai2022task,su2022one,wang2023improving,muennighoff2024generative}. Injecting instructions understanding abilities into retrievers are intuitively beneficial for modeling more nuanced semantics in natural language, such as intents, improving the alignment of query and groundtruth answer in reasoning tasks in the semantic space, especially for the ones that make less sense without specifying what the task is doing. Further, attaining more accurate understanding of queries from diverse tasks allows generalizing to retrieving correct answers for unseen tasks. Generalization achieved from task diversity and decoder architecture is known for generative tasks \citep{wei2021finetuned,wang2022language,wang2022super}, but more rigorous studies need to confirm the pattern for embedding models, especially for decoder models, which present the best potential on RAR-b.

\paragraph{Scaling laws observed} We observe a scaling behavior through the varied versions of models of same classes. The pattern is consistent for Instructor \citep{su2022one}, BGE \citep{xiao2023c} and OpenAI-Embedding-3. Models display a performance gain as their size increases. Since the level-1 datasets are more or less seen in Instructor's training through the SuperNI collection \citep{wang2022super}, its performance gain on these datasets is more relevant to a stronger fitting ability to training set due to larger model capacities, but not necessarily a stronger generalizability. On the other hand, the same pattern observed in BGE models \citep{xiao2023c} is more indicative of a stronger generalizability achieved with larger model sizes, because BGE models have not seem these tasks according to the technical report \citep{xiao2023c}. Although the finding is consistent with \citep{ni2022large} who find that larger-capacity models are more generalizable retrievers, it is unclear whether this comes from the difference in the pretraining data of the base model or the finetuning data when adapting to representation models, especially in our case of reasoning tasks. While Instructor and BGE are encoder models, the same trend observed in OpenAI v3 models confirms the scaling pattern for decoders, which are well-known to scale for generative tasks \citep{kaplan2020scaling,hoffmann2022training,wei2022emergent}, but less known for embedding tasks \citep{muennighoff2022sgpt}.

\paragraph{Distraction of Instructions} 
It is observed that instructions generally distract the retrievers' from focusing on the content, even for the I-DR models. However, this is more observed for encoder models like Instructor and BGE, and smaller variants of decoder models such as \texttt{OpenAI-embedding-3-small}. 

Latest instruction-aware decoder representational models \citep{wang2023improving,muennighoff2024generative} seem to start bridging this behavioral gap, pointing the promising direction of scaling decoder models for next-level representational abilities. Cohere's proprietary \texttt{Embed-English-v3.0} presents a strong gain brought by instructions. Considering its relatively early release time (Nov 02, 2023) compared to latest models that yield better absolute performance (OpenAI-v3, E5-Mistral, GritLM) on RAR-b, this behavior is surprising, and might be relevant to the efforts put in relevance feedback from LLMs \footnote{\url{https://txt.cohere.com/introducing-embed-v3/}}. However, no public information is available whether Cohere's current embedding model is an encoder or a decoder, and thus its strong instruction-following abilities can not add to our main narrative of the potentail for decoder representation models. 

For models that degrade with instructions, it is more severe in full-dataset retrieval settings compared to the Multiple-choice Retrieval setting as shown in later parts, intuitively because of a larger candidate pool to be distracted by. This phenomenon is most pronounced for older-generation embedding models such as Sentence Transformer \citep{reimers2019sentence} model \texttt{all-mpnet-base-v2}. Though sharing the same training set with \texttt{all-MiniLM-L6-v2} while both without training to follow instructions, we suspect it's more sensitive than the latter to instructions because of its larger capacity. The key takeaway here is, if a model is not trained to follow instructions, its retrieval performance might be largely distracted by instructions, hypothetically as the model capacity gets larger. 

With Sentence Transformer's earliest models being the pioneer in sentence embeddings in the contextualized LM's era, the field has now entered a stage where sentence emebedding models need to model more complex nuances in human language.

\paragraph{Dual-retrievers present decent performance, if not as late-interaction decision-makers.} In general, the nDCG@10 performance in Full setting is not as bad as the slightly-higher-than-chance performance in MCR setting as will be shown in later section (Table~\ref{tab: mcr-final}). The performance patterns are roughly on-par with how they would perform in pure topical-based IR datasets (e.g., the ones in BEIR \citep{thakur2021beir}). Judging from recall@10 (Appendix~\ref{appendix: full recall}), the percentages when correct answers can be extracted within top-10 documents are decent, except for SocialiQA. However, it is still far from meeting the capability of retrieving all candidates to assist later LLMs' decisions, which ideally require that at least the correct answer is in the top-n retrieved documents (if not ranked at top), i.e., recall@n $\approx$ 1.

\subsection{Does Reranker Help?}

\begin{figure}[ht]
    \centering
    \includegraphics[width=0.7\linewidth]{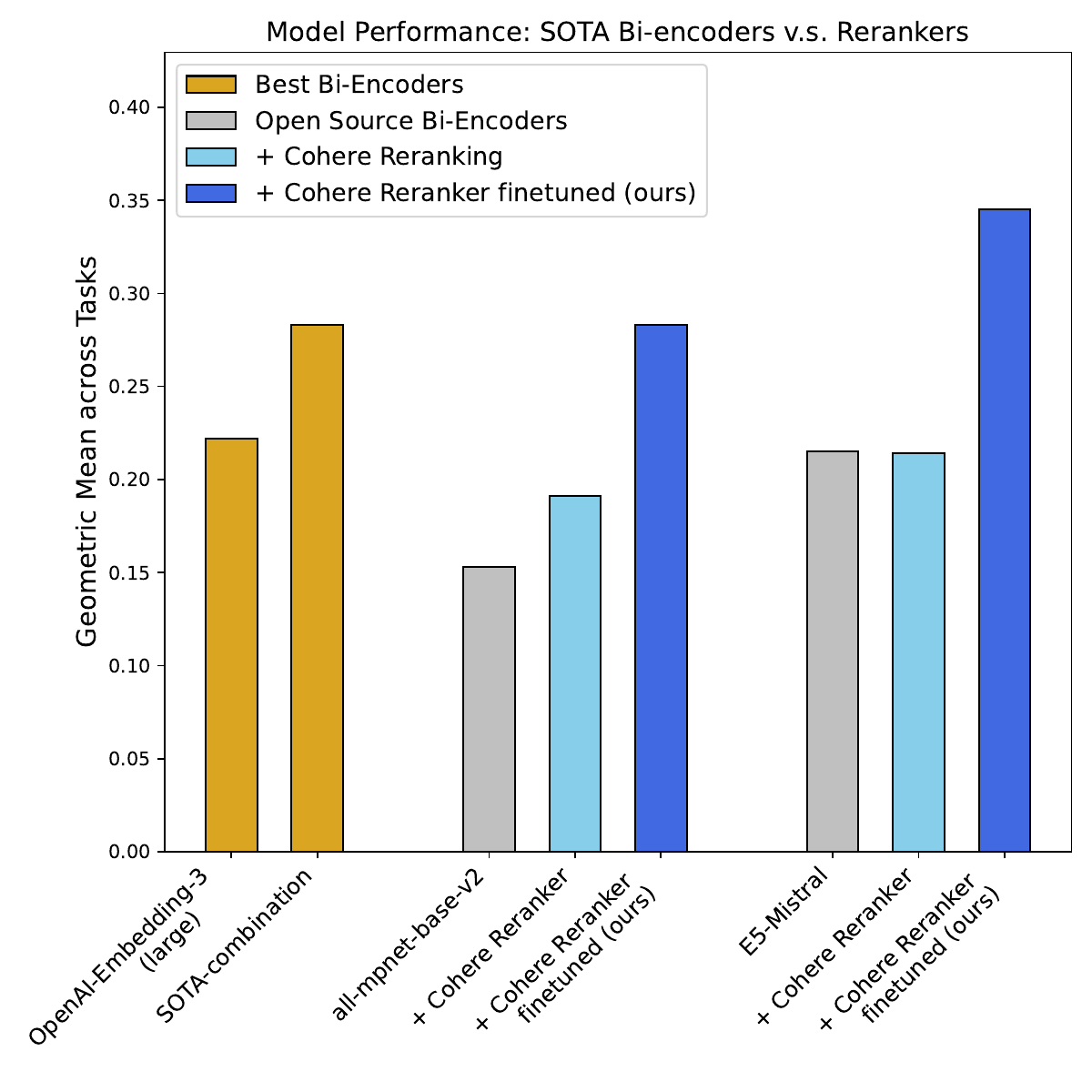}
    \caption{Does Reranking improve performance?}
    \label{fig: reranking performance}
\end{figure}

In a standard two-step retrieval pipeline, reranking models are utilized to further rerank the first-round results retrieved with bi-encoder embeddings. As we will show in the next sub-section through multiple-choice retrieval (Table~\ref{tab: mcr-final}), nowadays' reranker models are not trained to understand reasoning-level language, and are already outperformed by recent bi-encoders with decoder backbones trained with instruction tuning \citep{wang2023improving,muennighoff2024generative}. 

But can we fine-tune the reranker to improve performance on such tasks? Note that it is non-trivial to fine-tune a bi-encoder on each task: with varied sizes of each training set and the de-facto usage of InfoNCE loss \citep{oord2018representation}, we do not want to overfit a bi-encoder to an isotropic distribution for one reasoning task \citep{xiao2023isotropy}. And intuitively, cross-attention enables faster convergence to tasks with deeper language understanding requirements.

In this section, we finetune Cohere's proprietary \texttt{rerank-english-v2.0} model through API finetuning \footnote{https://dashboard.cohere.com/fine-tuning}, because of its well-configured infrastructure. For each task, we construct a training set using its original MCR-format dataset, with the groundtruth being the positive pair for each question, and all non-groundtruth candidates as hard negatives. We evaluate $\alpha$NLI, HellaSwag, PIQA, SocialiQA	and Winogrande to understand the pattern, and fine-tune a reranker model for each task. The average performance is outlined in Figure~\ref{fig: reranking performance} (without-instruction nDCG@10 presented), where we can easily achieve state-of-the-art performance by re-ranking the top 100 documents with a fine-tuned Cohere reranker. Note that the first-round retrieval result puts a large cap on the later re-ranking upperbound.

Although recent papers have confirmed generative models' potential as rerankers \citep{sun2023chatgpt,pradeep2023rankzephyr}, they are neither trivial to train nor cheap (yet) in inference. Considering the release time of Cohere's rerankers, we confirm using previous established cross-encoder frameworks is still the simple, robust and strong solution for such tasks. However, we envision decoder models to still be the future for scaling for complicated tasks, and position the effective training of decoder-based rerankers and resource-efficient inference of them as a valuable research direction.

\subsection{Multiple-choice Retrieval Setting} Table~\ref{tab: mcr-final} presents the results of the Multiple-choice Retrieval (MCR) setting, where each dataset has a baseline characterized by random predictions. The MCR setting provides an in-depth check of retrievers' nuanced understanding. Although we evaluate MCR with a separate implementation, we note that in essence, MCR is equivalent to checking whether the correct choice in ranked above all wrong choices in the Full setting.

The findings could be summarized as follows:

\paragraph{In-depth Understanding remains hard}

From Table~\ref{tab: mcr-final}, we can easily find that models struggle with Winogrande \citep{sakaguchi2021winogrande} and CSTS \citep{deshpande2023csts} at around chance level. Although some models perform good on Winogrande in the Full setting, we find that this performance might be more of a mirage. For example, Contriever can achieve an nDCG@10 of 0.471 on Winogrande without instruction in the Full setting (Table~\ref{tab: full-dataset ndcg@10}), outperforming even the OpenAI-Embedding-3-large model. However, its performance on MCR is stuck at chance level. We find that, the two candidate answers in Winogrande are both words that exist in the queries (e.g., entities), and it requires only finding these two words and ranked them at the top to get a good performance in the Full setting, making this task fall back on be a name entity matching task for the Full setting, without the models needing to develop a fine-grained understanding in distinguishing them. 

\begin{table}[!tp]\centering
\scalebox{0.72}{%
\begin{tabular}{lcccccc|c|cc|cc}\toprule
&\textbf{$\alpha$NLI} &\textbf{HellaSwag} &\textbf{PIQA} &\textbf{SIQA} &\textbf{WinoG} &\textbf{ARC-C} &\textbf{GeoMean-6} &\textbf{cSTS-E} &\textbf{cSTS-H} &\textbf{GeoMean} \\\midrule
\multicolumn{11}{l}{\textit{Without Instruction}} \\
\midrule
random baseline	&50.00	&25.00	&50.00&	33.33	&50.00	&25.00	&37.09	&50.00&	50.00	&39.97\\
contriever &51.37 &30.66 &57.51 &38.54 &49.80 &21.59 &39.42 &47.29 &43.11 &40.78 \\
mpnet &55.35 &34.98 &58.71 &41.50 &51.30 &27.90 &43.48 &49.65 &40.92 &43.87 \\
minilm &49.61 &32.20 &57.34 &39.41 &51.30 &24.06 &40.57 &47.38 &45.20 &41.92 \\
dragon &51.57 &31.40 &58.54 &39.30 &50.75 &24.15 &40.73 &45.29 &44.24 &41.70 \\
Instructor-base &55.94 &34.44 &61.15 &42.73 &49.49 &29.18 &44.01 &48.52 &51.66 &45.45 \\
Instructor-large &57.18 &38.48 &63.71 &46.78 &48.62 &31.48 &46.44 &47.47 &45.90 &46.50 \\
Instructor-XL &57.44 &39.80 &65.45 &43.86 &50.12 &31.48 &46.69 &49.83 &47.56 &47.18 \\
BGE-small &52.87 &32.56 &58.11 &41.20 &47.83 &23.29 &40.77 &44.42 &50.26 &42.30 \\
BGE-base &53.59 &33.58 &59.30 &42.37 &51.22 &25.09 &42.40 &48.08 &50.87 &44.06 \\
BGE-large &54.05 &34.89 &59.79 &42.73 &50.20 &26.19 &43.02 &49.30 &52.62 &44.87 \\
BGE-m3 &52.74 &33.23 &56.64 &45.45 &48.78 &24.83 &41.97 &43.72 &45.55 &42.62 \\
E5-Mistral &62.01 &42.40 &72.69 &50.72 &52.64 &48.38 &53.96 &55.58 &46.16 &53.11 \\
GritLM &62.27 &51.18 &72.85 &51.33 &53.12 &39.68 &54.12 &47.29 &46.16 &52.16 \\

Cohere-Embed-v3	&56.40	&34.59	&59.85	&50.00	&49.01	&25.60	&44.07	&46.77	&45.03	&44.52\\
OpenAI-3-small	&55.48 	&36.21 	&63.22 	&44.78 	&50.20 	&31.57 	&45.62 	&49.04 	&46.60 	&46.16 \\
OpenAI-3-large	&57.64 	&42.50 	&70.62 	&47.59 	&50.59 	&47.70 	&52.04 	&48.52 	&47.29 	&50.98 \\

\midrule
BGE-rerank-base &52.15 &31.86 &54.62 &40.58 &49.49 &21.93 &39.84 &47.38 &48.52 &41.73 \\
BGE-rerank-large &55.94 &35.38 &59.19 &44.47 &50.12 &26.79 &43.73 &48.08 &49.74 &44.97 \\
TART-Full &57.77 &36.01 &63.22 &48.98 &54.62 &34.04 &47.83 &45.55 &48.87 &47.67 \\
Cohere-rerank &57.25 &37.46 &64.04 &42.48 &50.28 &40.10 &47.69 &40.23 &38.13 &45.40 \\
\quad + finetuned &58.62 &50.57 &64.47 &57.32 &48.93 &43.43 &53.44 &55.67 &74.52 &55.99 \\
\midrule
\multicolumn{11}{l}{\textit{With Instruction}} \\
\midrule
random baseline	&50.00	&25.00	&50.00&	33.33	&50.00	&25.00	&37.09	&50.00&	50.00	&39.97\\
contriever &49.80 &30.87 &55.66 &38.95 &50.28 &22.53 &39.46 &49.74 &44.76 &41.26 \\
mpnet &54.11 &31.54 &56.26 &41.25 &51.46 &27.39 &42.12 &54.36 &40.05 &43.21 \\
minilm &50.26 &31.50 &56.86 &39.30 &49.57 &23.98 &40.17 &51.40 &45.81 &42.12 \\
dragon &51.57 &31.26 &57.24 &39.20 &51.54 &24.91 &40.85 &49.04 &44.50 &42.24 \\
Instructor-base &55.48 &33.82 &60.39 &42.63 &51.38 &29.44 &44.05 &48.60 &50.70 &45.39 \\
Instructor-large &57.77 &37.28 &63.66 &45.39 &49.96 &31.66 &46.29 &48.78 &44.76 &46.40 \\
Instructor-XL &57.05 &39.09 &64.53 &44.22 &50.67 &31.31 &46.49 &48.34 &46.07 &46.66 \\
BGE-small &52.15 &32.25 &56.69 &42.02 &49.57 &22.53 &40.59 &48.95 &52.09 &42.87 \\
BGE-base &53.39 &33.04 &57.34 &42.78 &50.75 &24.32 &41.81 &52.44 &48.52 &43.82 \\
BGE-large &54.37 &35.01 &59.30 &43.19 &49.09 &26.02 &42.90 &56.28 &51.66 &45.42 \\
BGE-m3 &54.31 &32.65 &56.53 &45.50 &48.78 &24.83 &42.04 &46.16 &47.03 &43.14 \\
E5-Mistral &64.30 &42.91 &73.78 &48.82 &53.67 &44.71 &53.65 &53.58 &46.95 &52.75 \\
GritLM &66.58 &52.16 &74.59 &56.04 &55.25 &53.50 &59.17 &47.03 &43.72 &55.36 \\
Cohere-Embed-v3	&56.53	&34.67	&59.09	&48.57	&50.04	&25.17	&43.82	&51.92	&48.25	&45.31\\
OpenAI-3-small	&54.90 	&34.02 	&58.92 	&45.14 	&49.49 	&30.55 	&44.25 	&52.36 	&48.60 	&45.72  \\
OpenAI-3-large	&58.42 	&40.30 	&64.96 	&50.00 	&51.85 	&45.99 	&51.30 	&55.24 	&46.60 	&51.16  \\

\midrule
BGE-rerank-base &52.35 &32.04 &55.50 &39.66 &49.41 &22.18 &39.92 &47.99 &53.23 &42.34 \\
BGE-rerank-large &53.92 &33.73 &56.96 &44.63 &50.12 &26.96 &42.92 &50.17 &51.66 &44.79 \\
TART-Full &57.96 &31.47 &63.11 &52.25 &55.56 &38.23 &48.35 &46.77 &55.06 &48.94 \\
Cohere-rerank &55.61 &35.91 &61.32 &42.37 &50.51 &40.10 &46.80 &41.19 &44.85 &45.82 \\
\quad + finetuned &58.29 &51.03 &64.36 &57.73 &49.64 &43.86 &53.73 &60.65 &75.39 &56.91 \\
\bottomrule
\end{tabular}}
\caption{MCR performance}\label{tab: mcr-final}
\end{table}

Multiple-choice retrieval settings serve as a proxy measurement of how well the correct reasoning can be found through representation matching, given a few selected possible answers already. From the MCR experiments, we can draw the conclusion that \textbf{current dual-dense retrievers fail to serve as late interaction reasoners}.

\paragraph{Generalization Gap} A special case in the models is Instructor, which is known to have seen the formats of the first 6 tasks through finetuning on Super-NI \citep{wang2022super}, enabling it to achieve a better performance on these tasks compared to traditional S-DR models. Yet, its performance on the two settings of CSTS that we construct is stuck at chance level. Without solid control experiments, we hypothesize that this pattern stems from the subpar generalizablity of encoder models. Although the pattern can currently be empirically supported by the better performance achieved by similarly instruction-tuned decoder models including E5-Mistral, GritLM and OpenAI models, the generalization gap between encoders and decoders has not yet been systematically investigated on information retrieval, and we position this as a valuable research topic.

\paragraph{Unsupervised Model is strong and robust to instructions.} Surprisingly, Contriever, as an unsupervised model only trained on different spans of text without labeled document pairs, serves as a strong baseline for all datasets. We speculate this pattern is due to that the formats of question-answer pair in reasoning datasets look alike the way that Contriever is trained on. For instance, a question and its answer in HellaSwag essentially compose a consecutive document, while being alike a span pair that Contriever is trained on. For this reason as well, Contriever is extremely robust to instructions, because a prefix instruction does not make the following question and answer looking less like a consecutive span in the document.

\section{Behavioral Analysis}

In this section, we present more in-depth understanding of retrievers' behaviors through the tasks that have multiple settings.

\subsection{Behavioral Analysis through the lens of Temporal Reasoning}
\label{subsec: temp reason}

With the 7 sub-settings we construct for TempReason (Table~\ref{tab: tempreason full retrieval}), we find that it provides a multi-faceted breakdown of retrievers' behaviors, providing understanding of their \textbf{parameterized knowledge}, \textbf{reading comprehension abilities}, and \textbf{robustness against irrelevant context}, and thus we provide an analysis of it in this sub-section.
TempReason-L1 (TR-1) in the original dataset concerns evaluating the built-in time perception of models, using time arithmetic questions. We evaluate this task with only one setting - the standalone setting (denoted as \textbf{Pure}). Query and groundtruth document in TR-1 looks like ``When is 7 months after Nov, 1998?" and ``June, 1999". 

For TempReason-L2 (TR-2) and TempReason-L3 (TR-3), we respectively construct three settings: \textbf{Pure}, \textbf{Fact}, and \textbf{Context+Fact}. These two levels of tasks evaluate question answering situated in time, providing us the understanding of the three aspects outlined above.

\paragraph{Parameterized Knowledge} \textit{Pure setting} here evaluates the built-in knowledge stored in retrievers. The question here can look like ``who is the president of United States in 2003?" (note that questions in TempReason are in a similar format but are much harder and niche than this), and the retrievers thus need to encode ``George W. Bush'' to be among having the most similar embedding. We can see the Pure setting evaluates  factual knowledge parameterized in retrievers without references. 

\paragraph{Reading Comprehension} For the \textit{Fact setting}, we concatenate the facts (a paragraph describing the answer to the question in different times - such as the history of US presidential tenures) after the question as query. Concretely, the query can look like: \texttt{Question: Who was the president of the United States in 2003? Facts: The US president from 2001 to 2009 was George W. Bush. Barack Obama's tenure began on January 20, 2009, and ended on January 20, 2017. Donald Trump began his presidency in 2017, and his tenure ended in 2021.} and ideally, \texttt{George W. Bush} has an embedding close to the above long passage. Note that this above example is for illustrating the concept with the same format as TempReason, and is not from the TempReason dataset, as TempReason is made of much less well-known people and facts, and thus more difficult. In the setting, we are essentially evaluating retrievers' reading comprehension abilities. Intuitively, the ``fact'' is laid out in the query and the embedding of the query is expected to be guided towards that of the groundtruth document, by the facts. Another way to describe this setting is that we let the retrievers play the role of LLM in a RAG system in this setting.

\paragraph{Robustness against Irrelevant Contexts}Lastly, for the Context+Fact setting, we further concatenate the context (the background knowledge of the entity in the question that is going to be asked) after the facts. On top of the Fact setting, the Context+Fact setting further reflects a model's robustness towards irrelevant information (since the answer can be extracted only from facts, but not from contexts), and encoding capabilities for texts of varied length ranges \citep{xiao2023length}. However, there potentially is unfair comparison in here because of the max sequence length each model is capable of encoding. For instance, OpenAI embedidng models are able to encode sequence up to 8192 tokens and are faced with more noise to be robust against. On the other hand, models with max sequence length of 512 need to deal with less noise. Therefore, the Context-Fact setting here is only for analysis purposes, and we do not average the scores from this setting into the average TR scores in the main table. For E5-Mistral and GritLM, we set its max sequence length to 8192 here to match with OpenAI models, for attaining a comparable understanding on the Context+Fact setting.

\begin{table}[!tp]\centering
\vspace{5pt}
\scalebox{0.74}{\begin{tabular}{lr|ccccccc|ccc}\toprule
\multicolumn{2}{c}{\textbf{Models →}} &\multicolumn{1}{c}{TR-1} &\multicolumn{3}{c}{TR-2} &\multicolumn{3}{c}{TR-3} & \multirow{2}{*}{avg.} & \multirow{2}{*}{avg. (fair)}\\\cmidrule{3-9}
\multicolumn{2}{c}{\textbf{Test Dataset ↓}} & pure &pure &fact &context+fact &pure &fact &context+fact \\\midrule

\multirow{2}{*}{Contriever} &w/o Inst. &0.019 &0.011 &0.227 &0.207 &0.078 &0.206&0.194&0.135&0.108 \\
&w/ Inst. &0.018 &0.009 &0.220 &0.205 &0.071 &0.208 &0.193 &0.132 &0.105\\
\midrule
\multirow{2}{*}{all-mpnet-base-v2} & w/o Inst. & 0.018 & 0.012 & 0.112 & 0.095 & 0.056 & 0.094 & 0.078 & 0.066 & 0.058 \\
& w/ Inst. & 0.015 & 0.010 & 0.073 & 0.076 & 0.052 & 0.071 & 0.067 & 0.052 & 0.044 \\
\multirow{2}{*}{all-MiniLM-L6-v2} & w/o Inst. & 0.015 & 0.005 & 0.176 & 0.100 & 0.063 & 0.141 & 0.087 & 0.084 & 0.080 \\
& w/ Inst. & 0.010 & 0.005 & 0.165 & 0.095 & 0.063 & 0.138 & 0.087 & 0.081 & 0.076 \\
\multirow{2}{*}{Dragon+} & w/o Inst. & 0.018 & 0.006 & 0.175 & 0.110 & 0.080 & 0.157 & 0.099 & 0.092 & 0.087 \\
& w/ Inst. & 0.015 & 0.006 & 0.161 & 0.110 & 0.075 & 0.148 & 0.095 & 0.087 & 0.081 \\
\multirow{2}{*}{bge-m3} & w/o Inst. & 0.010 & 0.007 & 0.332 & 0.223 & 0.053 & 0.301 & 0.192 & 0.160 & 0.140 \\
& w/ Inst. & 0.008 & 0.006 & 0.350 & 0.263 & 0.070 & 0.325 & 0.230 & 0.179 & 0.152 \\
\midrule
\multirow{2}{*}{Instructor-base} & w/o Inst. & 0.007 & 0.006 & 0.109 & 0.066 & 0.055 & 0.120 & 0.076 & 0.063 & 0.059 \\
& w/ Inst. & 0.007 & 0.005 & 0.091 & 0.064 & 0.055 & 0.100 & 0.073 & 0.056 & 0.052 \\
\multirow{2}{*}{Instructor-large} & w/o Inst. & 0.007 & 0.011 & 0.123 & 0.081 & 0.069 & 0.133 & 0.084 & 0.073 & 0.068 \\
& w/ Inst. & 0.007 & 0.011 & 0.104 & 0.070 & 0.072 & 0.117 & 0.079 & 0.066 & 0.062 \\
\multirow{2}{*}{Instructor-XL} & w/o Inst. & 0.006 & 0.013 & 0.171 & 0.099 & 0.074 & 0.156 & 0.093 & 0.088 & 0.084 \\
& w/ Inst. & 0.008 & 0.013 & 0.144 & 0.089 & 0.077 & 0.137 & 0.088 & 0.079 & 0.076 \\
\multirow{2}{*}{bge-small} & w/o Inst. & 0.014 & 0.010 & 0.176 & 0.070 & 0.048 & 0.139 & 0.059 & 0.074 & 0.077 \\
& w/ Inst. & 0.013 & 0.011 & 0.167 & 0.085 & 0.046 & 0.128 & 0.074 & 0.075 & 0.073 \\
\multirow{2}{*}{bge-base} & w/o Inst. & 0.011 & 0.013 & 0.172 & 0.099 & 0.052 & 0.134 & 0.082 & 0.080 & 0.076 \\
& w/ Inst. & 0.008 & 0.013 & 0.166 & 0.111 & 0.051 & 0.127 & 0.092 & 0.081 & 0.073 \\
\multirow{2}{*}{bge-large} & w/o Inst. & 0.015 & 0.024 & 0.242 & 0.177 & 0.067 & 0.206 & 0.146 & 0.125 & 0.111 \\
& w/ Inst. & 0.012 & 0.021 & 0.212 & 0.158 & 0.060 & 0.176 & 0.126 & 0.109 & 0.096 \\
\multirow{2}{*}{E5-Mistral}	& w/o Inst.	&0.030	&0.093	&0.356	&0.247	&0.144	&0.304	&0.213	&0.198	&0.185\\
&w/ Inst.	&0.033	&0.092	&0.369	&0.281	&0.143	&0.302	&0.231	&0.207	&0.188\\
\multirow{2}{*}{GritLM}	& w/o Inst.	&0.025	&0.090	&0.482	&0.289	&0.125	&0.340	&0.225	&0.225	&0.212\\
&w/ Inst.	&0.072	&0.112	&0.576	&0.335	&0.141	&0.439	&0.261	&0.277	&0.268\\
\midrule
\multirow{2}{*}{Cohere-Embed-v3} & w/o Inst. & 0.015 & 0.019 & 0.359 & 0.197 & 0.085 & 0.275 & 0.162 & 0.159 & 0.151 \\
& w/ Inst. & 0.014 & 0.024 & 0.405 & 0.249 & 0.075 & 0.339 & 0.213 & 0.188 & 0.171 \\
\multirow{2}{*}{OpenAI-ada-2} & w/o Inst. & 0.017 & 0.026 & 0.199 & 0.163 & 0.076 & 0.180 & 0.148 & 0.116 & 0.100 \\
& w/ Inst. & 0.014 & 0.024 & 0.194 & 0.161 & 0.073 & 0.176 & 0.147 & 0.113 & 0.096 \\
\multirow{2}{*}{OpenAI-3-small} & w/o Inst. & 0.023 & 0.028 & 0.257 & 0.165 & 0.098 & 0.221 & 0.152 & 0.135 & 0.125 \\
& w/ Inst. & 0.023 & 0.032 & 0.263 & 0.165 & 0.100 & 0.227 & 0.152 & 0.137 & 0.129 \\
\multirow{2}{*}{OpenAI-3-large} & w/o Inst. & 0.021 & 0.103 & 0.286 & 0.187 & 0.153 & 0.255 & 0.165 & 0.167 & 0.164 \\
& w/ Inst. & 0.021 & 0.110 & 0.398 & 0.272 & 0.155 & 0.370 & 0.233 & 0.223 & 0.211 \\
\bottomrule
\end{tabular}}
\caption{TempReason all sub-task Full-Retrieval nDCG@10 Results. For TempReason-L2 and TempReason-L3, we construct 3 settings: Pure, Fact, and Context+Fact. The pure setting reflects the standalone knowledge parameterized into retrieval models. The Fact setting reflects the reading comprehension abilities of retrieval models. And the Context+Fact setting is an indicator of retrievers' reading comprehension abilities subtracted by their vulnerability against irrelevant contexts.}\label{tab: tempreason full retrieval}
\end{table}

\subsection{Behavioral Analysis through the lens of Code Retrieval}

Another area that we can get a more fine-grained understanding about retrievers' behaviors is through code retrieval, given the extensive settings we explored in the construction of the final pooled evaluation dataset.

\paragraph{Hard Negatives across Programming Languages}

The translated nature of HumanEvalPack \citep{muennighoff2023octopack} grants us the convenience to inspect models' familiarity with the same content of different programming languages, and when pooling them together, to what extent do the hard negatives come from the failure of distinguishing across languages through instructions (e.g., for entries that require ``retrieve a Python snippet'', a Javascript snippet of same content is instead ranked above the groundtruth Python snippet).

\begin{figure}
    \centering
    \includegraphics[width=1\linewidth]{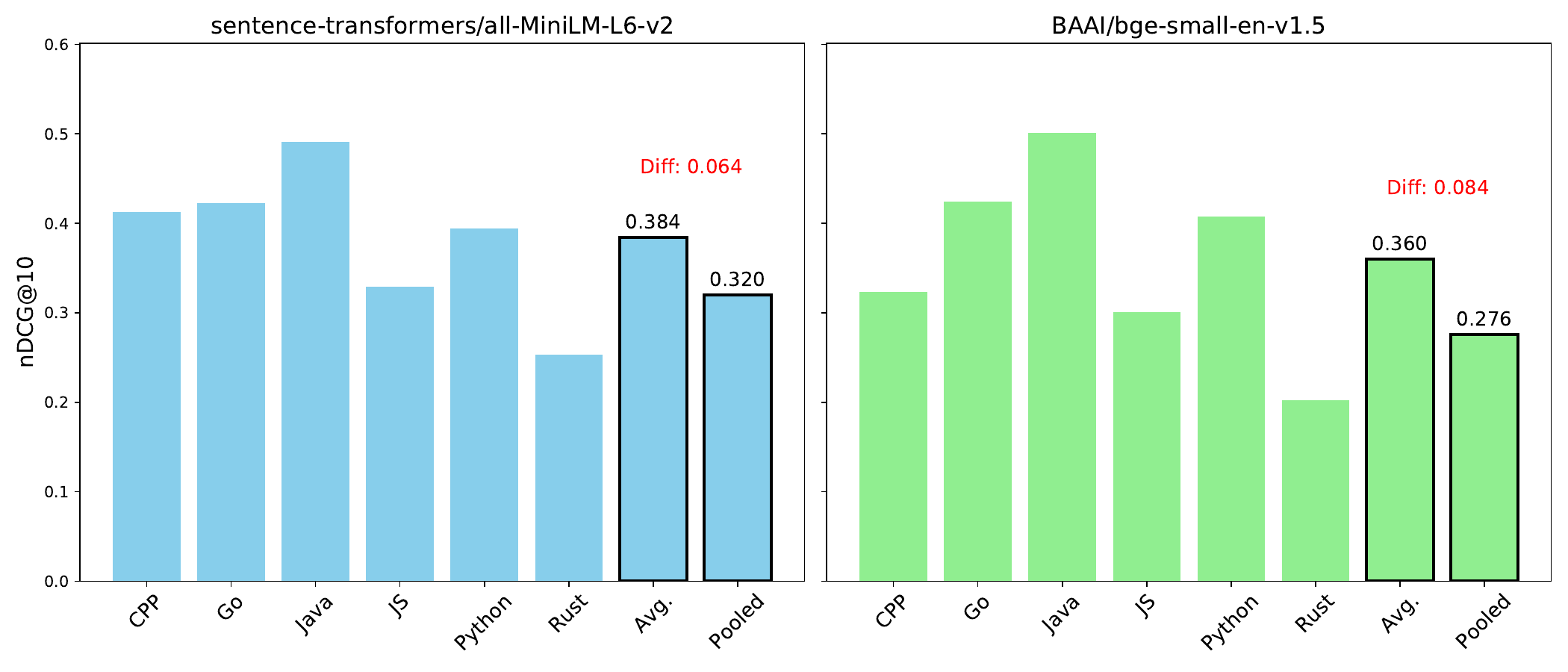}
    \caption{A glimpse of distraction from hard negatives of same content across languages.}
    \label{fig: code ablation1}
\end{figure}

Here, we inspect the results of 7 settings, respectively with only queries from each of the 6 languages, and the pooled queries from all languages. For each setting, we enlarge the corpus with the same 200k documents sampled from CodeSearchNet and 100k documents from TinyCode, aligning with our final setting. However, we see a performance gap between averaging individual settings and the pooled setting, meaning that for a lot of queries, the ``semi-groundtruth'' is placed over the groundtruth, indicating the model's failure of perceiving about programming languages through instructions.

Figure~\ref{fig: code ablation1} depicts the pattern. If the queries are not distracted by code of same content from different languages (i.e., same content from different languages is ranked above the groundtruth), the average performance of the single-language settings should actually match with the performance of the pooled setting, because in the pooled setting, the nDCG@10 is essentially nDCG@10 performance averaged across different languages.

\paragraph{Entity Shortcut}
\label{sec: ablation entity shortcut}

\begin{table}[!htp]\centering
\scriptsize
\begin{tabular}{lrcc|c}\toprule
\multirow{2}{*}{Settings}& &\multicolumn{2}{c}{HumanEvalPack}&TinyCode\\
\cmidrule{3-5}
& format $\rightarrow$ &Original &Instruct & -  \\\midrule
\multirow{2}{*}{Contriever} &w/o Inst. &0.051 &0.203 &0.056 \\
&w/ Inst. &0.048 &0.228 &0.044 \\
\multirow{2}{*}{all-mpnet-base-v2} &w/o Inst. &0.419 &0.897 &0.142 \\
&w/ Inst. &0.408 &0.867 &0.123 \\
\multirow{2}{*}{Instructor-XL} &w/o Inst. &0.396 &0.920 &0.209 \\
&w/ Inst. &0.374 &0.915 &0.240 \\
\bottomrule
\end{tabular}
\caption{Left: Performance difference between Original and Instruct setting of HumanEvalPack. Right: Performance of TinyCode, providing a glimpse of model behaviors on mixture of natural language and code.}\label{tab: code-ablation2}
\end{table}

Another property we present is the Entity Shortcut. In Table~\ref{tab: code-ablation2} (left), we present the results of two settings we construct for evaluating HumanEvalPack. For the original setting, we simply follow the continuation setting, where the query is the import, function name and the docstring of the function, and the aim is to retrieve the rest of the code. We construct another setting, the ``instruct'' setting, where we would like to retrieve the complete code (the concatenation of the query and the groundtruth in the ``original'' setting) with a self-contained instruction.

With the Instruct setting, we see a huge jump of nDCG@10 from the original setting (e.g., 0.374 $\rightarrow$ 0.915 for Instructor-XL), indicating that the easiness of this setting for the model. An example for this setting is given in Appendix~\ref{appendix: entity matching shortcut}, demonstrating the entity matching shortcut. Although this setting forms a more self-contained scenario that would happen in real-life question answering, the function declaration presents in both the query and the document, enabling the retrieval to rely solely on the function declaration matching. We speculate this pattern to be highly relevant to the ``entourage effect" towards dominant tokens (function declaration in this case) after contrastive learning found in \citep{xiao2023isotropy}. Due to this reason, we opt for the original format in the final evaluation.

\paragraph{Natural v.s. Symbolic Language}

We further conduct an experiment by constructing the queries and corpus with only TinyCode, a synthetic dataset of code question answering. This dataset provides the natural mixture of natural language and code. We sample 15k queries and 150k documents including the groundtruth to form the corpus. The result is shown in Table~\ref{tab: code-ablation2} (right). Although hard to construct controlled experiments, it is seen that models generally perform less well on the mixture of natural language and code.

\section{Conclusion}

In this paper, we underscored the critical need of evaluating retrievers' next-level language understanding abilities, moving beyond the conventional scope of semantic textual similarity and topical matching. We introduced the RAR framework, a novel problem that casts reasoning tasks as retrieval tasks. Through extensive evaluation, we revealed retrievers' behaviors in understanding reasoning-level language and depicted the behavioral discrepancies between retrievers and LLMs, suggesting promising directions in future research in retrieval-augmented generation. We assert that representation models excelling in RAR-b are well-suited to enhance LLMs within RAG systems, positioning RAR-b as a vital benchmark for tracking advancements in the field.

\section*{Acknowledgement}

This study was partially supported by compute credits from the Cohere For AI Research Grant, which is designed to support academic partners conducting research with the goal of releasing scientific artifacts and data for good projects.

\bibliography{iclr2021_conference}
\bibliographystyle{iclr2021_conference}

\appendix
\section{Dataset Format}
\label{appendix: dataset format}

{\scriptsize \begin{longtable}{ p{.12\textwidth} | p{.25\textwidth} | p{.3\textwidth} | p{.2\textwidth}} \toprule
Dataset&Instruction &Query &Doc \\\midrule
$\alpha$NLI &Given the following start and end of a story, retrieve a possible reason that leads to the end. &Start: Ron started his new job as a landscaper today. End: Ron is immediately fired for insubordination. &Ron ignores his bosses's orders and called him an idiot. \\
\midrule
HellaSwag &Given the following unfinished context, retrieve the most plausible ending to finish it. &A man is sitting on a roof. He &starts pulling up roofing on a roof. \\
\midrule
PIQA &Given the following goal, retrieve a possible solution. &How do I ready a guinea pig cage for it's new occupants? &Provide the guinea pig with a cage full of a few inches of bedding made of ripped paper strips, you will also need to supply it with a water bottle and a food dish. \\
\midrule
SocialiQA &Given the following context and question, retrieve the correct answer. &Context: Tracy didn't go home that evening and resisted Riley's attacks. Question: What does Tracy need to do before this? &Find somewhere to go \\
\midrule
Quail & Given the following context and question, retrieve the correct answer. & Context: Candy watched the bearded man drive his silver BMW into the convenience store parking lot and pull around to the side, near the back corner of the building. There were plenty of open slots in the front, so she figured the guy was there for something other than a bag of chips and a coke.\textbackslash nAchilly breeze blew up her mini-skirt and she shivered. She pressed her legs together tightly to generate some heat. The knee-high boots protected her feet and calves, but her butt was freezing off.\textbackslash nShe wrote down the license number as she circled around to the side of the expensive vehicle. He'll have a big wad of cash, she thought.\textbackslash nLarry Luzor had just stepped out of the car, when she said, "Nice car, Honey."\textbackslash n"Uh, thanks."\textbackslash n"I'm Candy. You got a sweet tooth tonight?" He gave her the once over. Her black hair framed a pretty, young-looking face. The low-cut blouse left little to the imagination, barely hiding her nipples. She was average height, but the high heel boots elevated her to about 5'8". The long legs were very nice.\textbackslash nLarry had never used a prostitute. He'd always thought of it as revolting. The idea of having sex with a woman who'd been with hundreds of men did not appeal to him.\textbackslash nBut this didn't seem like a typical hooker. She seemed too clean--almost pure. But of course, she wasn't. He knew she had to be just as skanky as the rest of them. Still--if he hadn't been in the middle of something important he might have been more than willing to buy what she was selling.\textbackslash n"So, what do you say? Want to get it on?" She smiled seductively.\textbackslash nHe was impressed that she had all her teeth, and that they looked white.\textbackslash n"How can I resist?" He grinned at her and winked.& about 10 minutes \\
\midrule
ARC-Challenge &Retrieve the answer to the question. &An astronomer observes that a planet rotates faster after a meteorite impact. Which is the most likely effect of this increase in rotation?" &Planetary days will become shorter. \\
\midrule
Winogrande &Given the following sentence, retrieve an appropriate answer to fill in the missing underscored part. &Sentence: Sarah was a much better surgeon than Maria so \_ always got the easier cases.. &Maria \\
\midrule
TempReason-L1 &Given the following question about time, retrieve the correct answer. &What is the time 6 year and 4 month after Nov, 1185 &Mar, 1192\\
\midrule
TempReason-L2 pure &Given the following question, retrieve the correct answer. &Which employer did Jaroslav Pelikan work for in Jan, 1948? &Valparaiso University \\
\midrule
TempReason-L2 fact &Given the following question and facts, retrieve the correct answer. &Question: Which employer did Jaroslav Pelikan work for in Jan, 1948? Facts: Jaroslav Pelikan works for Concordia Seminary from Jan, 1949 to Jan, 1953. \textbackslash n Jaroslav Pelikan works for University of Chicago from Jan, 1953 to Jan, 1962. \textbackslash n Jaroslav Pelikan works for Valparaiso University from Jan, 1946 to Jan, 1949. \textbackslash n Jaroslav Pelikan works for Yale University from Jan, 1962 to Jan, 1962. &Valparaiso University \\
\midrule
TempReason-L2 context &Given the following question, facts and contexts, retrieve the correct answer. &Question: Which employer did Jaroslav Pelikan work for in Jan, 1948? Facts: Jaroslav Pelikan works for Concordia Seminary from Jan, 1949 to Jan, 1953. \textbackslash n Jaroslav Pelikan works for University of Chicago from Jan, 1953 to Jan, 1962. \textbackslash n Jaroslav Pelikan works for Valparaiso University from Jan, 1946 to Jan, 1949. \textbackslash n Jaroslav Pelikan works for Yale University from Jan, 1962 to Jan, 1962. Context: Jaroslav PelikanJaroslav Jan Pelikan Jr. (December 17, 1923 \textbackslash u2013 May 13, 2006) was an American scholar of the history of Christianity, Christian theology, and medieval intellectual history at Yale University.Jaroslav Jan Pelikan Jr. was born on December 17, 1923, in Akron, Ohio, to a Slovak father Jaroslav Jan Pelikan Sr. and Slovak mother Anna Buzekova Pelikan from \textbackslash u0160id in Serbia. His father was pastor of .... \textbf{\{context is too long and omitted for brevity\}} &Valparaiso University\\
\midrule
TempReason-L3 pure &Given the following question, retrieve the correct answer. &Which employer did Jaroslav Pelikan work for before Concordia Seminary? &Valparaiso University \\
\midrule
TempReason-L3 fact &Given the following question and facts, retrieve the correct answer. &Question: Which employer did Jaroslav Pelikan work for before Concordia Seminary? Facts: Jaroslav Pelikan works for Yale University from Jan, 1962 to Jan, 1962. \textbackslash nJaroslav Pelikan works for University of Chicago from Jan, 1953 to Jan, 1962. \textbackslash nJaroslav Pelikan works for Valparaiso University from Jan, 1946 to Jan, 1949. \textbackslash nJaroslav Pelikan works for Concordia Seminary from Jan, 1949 to Jan, 1953. &Valparaiso University \\
\midrule
TempReason-L3 context &Given the following question, facts and contexts, retrieve the correct answer. &Question: Which employer did Jaroslav Pelikan work for before Concordia Seminary? Facts: Jaroslav Pelikan works for Yale University from Jan, 1962 to Jan, 1962. \textbackslash nJaroslav Pelikan works for University of Chicago from Jan, 1953 to Jan, 1962. \textbackslash nJaroslav Pelikan works for Valparaiso University from Jan, 1946 to Jan, 1949. \textbackslash nJaroslav Pelikan works for Concordia Seminary from Jan, 1949 to Jan, 1953. Context: Jaroslav PelikanJaroslav Jan Pelikan Jr. (December 17, 1923 \textbackslash u2013 May 13, 2006) was an American scholar of the history of Christianity, Christian theology, and medieval intellectual history at Yale University.Jaroslav Jan Pelikan Jr. was born on December 17, 1923, in Akron, Ohio, to a Slovak father Jaroslav Jan Pelikan Sr. and Slovak mother Anna Buzekova Pelikan from \textbackslash u0160id in Serbia. His father was pastor of .... \textbf{\{context is too long and omitted for brevity\}}&Valparaiso University \\
\midrule
SpartQA &Given the following spatial reasoning question, retrieve the right answer. &We have three blocks. Lets call them A, B and C. Block B is below A. Block A is below C. Block A contains a medium yellow square. Block B has two medium blue squares. Medium blue square number one is touching the bottom edge of this block. Medium blue square number two is below a medium yellow square. Medium blue square number one is below the square which is below the medium yellow square. It is below the medium yellow square. Block C contains one medium black square. What is below the black thing? a medium yellow square that is in block A or a medium yellow square that is in block B? &both of them \\
\midrule
Math-pooled & Retrieve the answer for the following math problem. & \textbf{\{Omitted for brevity\}} & \textbf{\{Omitted for brevity\}}\\
\midrule
HumanEvalPack-MBPP &Retrieve the answer for the following coding problem. &Finish the following code based on the docstring: \textbf{\{Omitted for brevity\}} & \textbf{\{Omitted for brevity\}} \\
\midrule
CSTS-easy & Retrieve an aspect and a sentence which are similar to the following sentence.	& 3 people wearing festive costumes and feathers dancing in a parade . & In terms of "The number of persons" retrieve a sentence similar to the following. Three uniquely dressed women dancing in a parade .\\
\midrule
CSTS-hard&Retrieve a condition under which the following two sentences are similar.&Sentence1: 3 people wearing festive costumes and feathers dancing in a parade .; Sentence2: Three uniquely dressed women dancing in a parade .&The number of persons.\\

\bottomrule
\end{longtable}}



\section{Potential Sparse Annotation Problem of C-STS}
\label{appendix: csts sparse annotation}

CSTS concerns aspect-aware similarity perception between sentence pairs. For each unique sentence pair in C-STS, two conditions are given, yielding two annotated similarity scores ranging from 1-5. Based on the nature of the dataset, we are only able to construct it into multiple-choice setting, evaluating the retrieval of the condition out of the 2 that provides a higher similarity, but can not construct it into Full-dataset retrieval setting because of the potential sparse annotation problem \citep{thakur2021beir} - a condition, based on which two sentences are more similar, than based on the other condition, does not make this condition the most suitable condition out of the corpus. 

\section{Illustration of the Entity Matching Shortcut}
\label{appendix: entity matching shortcut}

In Section~\ref{sec: ablation entity shortcut}, we presented the high performance of the Instruct setting of HumanEvalPack, which is largely due to the entity matching shortcut. We provide an example of a typical query-groudtruth pair in this setting as follows:

Query in ``Instruct'' setting:

\begin{minipage}{\textwidth}
\small\texttt{Write a Python function `has\_close\_elements(numbers: List[float], threshold: float) -> bool` to solve the following problem:\\Check if in given list of numbers, are any two numbers closer to each other than\\given threshold.\\>>> has\_close\_elements([1.0, 2.0, 3.0], 0.5)\\False\\>>> has\_close\_elements([1.0, 2.8, 3.0, 4.0, 5.0, 2.0], 0.3)\\True}
\end{minipage}

Groundtruth in ``Instruct'' setting:

\begin{minipage}{\linewidth}
\small\begin{verbatim}
from typing import List


def has_close_elements(numbers: List[float], threshold: float) -> bool:
    for idx, elem in enumerate(numbers):
        for idx2, elem2 in enumerate(numbers):
            if idx != idx2:
                distance = abs(elem - elem2)
                if distance < threshold:
                    return True

    return False
\end{verbatim}
\end{minipage}

With function declaration presenting in both query and groundtruth document, the document can easily be retrieved with the overlapping part perceived by the retrievers.

\section{Full setting Recall@10}
\label{appendix: full recall}

Table~\ref{tab: full-dataset recall@10} presents the recall@10 performance of the Full retrieval setting, providing a more straightforward measure of the average empirical probability that the groundtruth is retrieved as the top-10 documents.

\begin{table}[h]\centering

\scalebox{0.62}{%
\begin{tabular}{lr|rcccccccccc|cc}\toprule
\multicolumn{2}{l}{Model $\downarrow$ Dataset  $\rightarrow$} & $\alpha$NLI &HellaSwag &PIQA &SIQA &Quail &ARC-C &WinoG  &TR &SpartQA &Math &Code&Geo. Mean\\\midrule
\multirow{2}{*}{Contriever} &w/o Inst. &0.425 &0.221 &0.387 &0.019 &0.097 &0.161 &0.717 &0.210 &0.196 &0.357 &0.142 &0.196 \\
&w/ Inst. &0.371 &0.271 &0.335 &0.016 &0.096 &0.147 &0.463 &0.202 &0.184 &0.258 &0.110 &0.171 \\

\midrule

\multirow{2}{*}{all-mpnet-base-v2} &w/o Inst. &0.310 &0.396 &0.439 &0.041 &0.092 &0.207 &0.386 &0.112 &0.006 &0.786 &0.698 &0.178 \\
&w/ Inst. &0.039 &0.207 &0.426 &0.021 &0.081 &0.179 &0.195 &0.087 &0.026 &0.765 &0.654 &0.133 \\
\multirow{2}{*}{all-MiniLM-L6-v2} &w/o Inst. &0.393 &0.373 &0.392 &0.024 &0.076 &0.170 &0.770 &0.151 &0.038 &0.749 &0.588 &0.210 \\
&w/ Inst. &0.228 &0.321 &0.383 &0.029 &0.076 &0.162 &0.401 &0.143 &0.014 &0.700 &0.573 &0.169 \\
\multirow{2}{*}{Dragon+} &w/o Inst. &0.425 &0.421 &0.412 &0.028 &0.081 &0.166 &0.948 &0.167 &0.182 &0.517 &0.251 &0.232 \\
&w/ Inst. &0.355 &0.363 &0.390 &0.025 &0.079 &0.154 &0.890 &0.156 &0.201 &0.423 &0.197 &0.210 \\
\multirow{2}{*}{bge-m3} &w/o Inst. &0.350 &0.392 &0.354 &0.079 &0.136 &0.158 &0.692 &0.202 &0.148 &0.764 &0.546 &0.276 \\
&w/ Inst. &0.352 &0.391 &0.297 &0.076 &0.128 &0.158 &0.607 &0.215 &0.136 &0.722 &0.560 &0.265 \\

\midrule
\multirow{2}{*}{Instructor-base} &w/o Inst. &0.349 &0.403 &0.424 &0.040 &0.069 &0.181 &0.538 &0.113 &0.074 &0.671 &0.570 &0.217 \\
&w/ Inst. &0.298 &0.364 &0.392 &0.042 &0.071 &0.177 &0.311 &0.101 &0.120 &0.665 &0.542 &0.207 \\
\multirow{2}{*}{Instructor-large} &w/o Inst. &0.354 &0.437 &0.497 &0.059 &0.097 &0.221 &0.510 &0.136 &0.032 &0.788 &0.728 &0.236 \\
&w/ Inst. &0.335 &0.402 &0.442 &0.056 &0.083 &0.192 &0.425 &0.124 &0.066 &0.746 &0.696 &0.230 \\
\multirow{2}{*}{Instructor-XL} &w/o Inst. &0.455 &0.477 &0.527 &0.061 &0.111 &0.247 &0.873 &0.157 &0.008 &0.678 &0.658 &0.231 \\
&w/ Inst. &0.400 &0.449 &0.470 &0.075 &0.102 &0.204 &0.594 &0.143 &0.045 &0.655 &0.659 &0.248 \\
\multirow{2}{*}{bge-small} &w/o Inst. &0.175 &0.390 &0.360 &0.016 &0.033 &0.165 &0.228 &0.144 &0.078 &0.543 &0.573 &0.160 \\
&w/ Inst. &0.024 &0.357 &0.320 &0.018 &0.039 &0.142 &0.122 &0.138 &0.058 &0.548 &0.567 &0.121 \\
\multirow{2}{*}{bge-base} &w/o Inst. &0.170 &0.403 &0.389 &0.022 &0.028 &0.174 &0.332 &0.146 &0.070 &0.565 &0.615 &0.170 \\
&w/ Inst. &0.072 &0.364 &0.353 &0.007 &0.025 &0.161 &0.242 &0.141 &0.058 &0.540 &0.619 &0.130 \\
\multirow{2}{*}{bge-large} &w/o Inst. &0.195 &0.428 &0.418 &0.024 &0.035 &0.177 &0.419 &0.203 &0.061 &0.666 &0.637 &0.190 \\
&w/ Inst. &0.022 &0.392 &0.363 &0.012 &0.052 &0.161 &0.211 &0.180 &0.048 &0.587 &0.608 &0.131 \\
\multirow{2}{*}{E5-Mistral} &w/o Inst. &0.268 &0.472 &0.478 &0.088 &0.128 &0.316 &0.696 &0.327 &0.185 &0.848 &0.954 &0.341 \\
&w/ Inst. &0.377 &0.508 &0.561 &0.090 &0.144 &0.276 &0.678 &0.333 &0.196 &0.816 &0.945 &0.360 \\
\multirow{2}{*}{GritLM} &w/o Inst. &0.410 &0.501 &0.511 &0.091 &0.159 &0.271 &0.852 &0.374 &0.030 &0.888 &0.952 &0.318 \\
&w/ Inst. &0.464 &0.542 &0.618 &0.117 &0.207 &0.411 &0.840 &0.434 &0.189 &0.881 &0.960 &0.429 \\

\midrule[0.5pt]\midrule[0.5pt]

\multirow{2}{*}{Cohere-Embed-v3} &w/o Inst. &0.223 &0.400 &0.423 &0.073 &0.082 &0.180 &0.886 &0.275 &0.080 &0.790 &0.739 &0.266 \\
&w/ Inst. &0.272 &0.441 &0.417 &0.082 &0.140 &0.182 &0.939 &0.298 &0.074 &0.787 &0.737 &0.291 \\
\multirow{2}{*}{OpenAI-ada-2} &w/o Inst. &0.357 &0.441 &0.465 &0.051 &0.110 &0.238 &0.342 &0.188 &0.087 &0.809 &0.914 &0.262 \\
&w/ Inst. &0.178 &0.381 &0.378 &0.046 &0.108 &0.213 &0.199 &0.181 &0.094 &0.756 &0.903 &0.221 \\
\multirow{2}{*}{OpenAI-3-small} &w/o Inst. &0.419 &0.464 &0.498 &0.048 &0.113 &0.259 &0.543 &0.237 &0.137 &0.783 &0.880 &0.298 \\
&w/ Inst. &0.311 &0.415 &0.460 &0.048 &0.120 &0.235 &0.441 &0.242 &0.079 &0.718 &0.879 &0.264 \\
\multirow{2}{*}{OpenAI-3-large} &w/o Inst. &0.504 &0.498 &0.596 &0.054 &0.189 &0.388 &0.511 &0.306 &0.167 &0.940 &0.966 &0.362 \\
&w/ Inst. &0.467 &0.468 &0.559 &0.085 &0.233 &0.358 &0.577 &0.363 &0.160 &0.921 &0.964 &0.383 \\
\bottomrule
\end{tabular}}
\caption{Full-dataset Retrieval (recall@10 performance)}\label{tab: full-dataset recall@10}
\end{table}

\end{document}